\documentclass[letterpaper, 10 pt, conference]{ieeeconf}  % Comment this line out if you need a4paper

\IEEEoverridecommandlockouts                              % This command is only needed if 
\usepackage{graphicx}
\usepackage{times}
\usepackage{epsfig}
\usepackage{amsmath}
\usepackage{algorithm}
\usepackage{amssymb}
\usepackage{multirow}
\usepackage{mathtools,xparse}
\usepackage{siunitx}
\usepackage{stmaryrd}
\usepackage{hyperref}
\usepackage{amsmath}
\usepackage[noend]{algpseudocode}
\usepackage{placeins}
\usepackage{mdframed}
\usepackage{amsmath,amssymb} % define this before the line numbering.
\usepackage{color}
\usepackage{subcaption}
\usepackage[table]{xcolor}

\algnewcommand\algorithmicforeach{\textbf{for each}}
\algdef{S}[FOR]{ForEach}[1]{\algorithmicforeach\ #1\ \algorithmicdo}

\newcommand{\trans}{\text{T}}
\newcommand{\matr}[1]{\mathbf{#1}}     % ISO complying version

\newcommand{\Proj}{\matr{P}}

\newcommand{\Aff}{\matr{A}}
\newcommand{\Fund}{\matr{F}}
\newcommand{\Ess}{\matr{E}}
\newcommand{\Intrinsic}{\matr{C}}
\newcommand{\Point}{\matr{p}}
\newcommand{\Qoint}{\matr{q}}

\title{\LARGE \bf
Relative planar motion for vehicle-mounted cameras from a single affine correspondence
}

\author{Levente Hajder$^{1}$ and Daniel Barath$^{2}$% <-this % stops a space
\thanks{*Levente Hajder was supported by the project EFOP-3.6.3-VEKOP-16-2017-00001: Talent Management in Autonomous Vehicle Control Technologies, by the Hungarian Government and co-financed by the European Social Fund. His work was also supported by Thematic Excellence Programme, Industry and Digitization Subprogramme, NRDI Office, 2019. Daniel Barath was supported by the Hungarian Scientific Research Fund (No.\ NKFIH OTKA KH-126513 and K-120499) and by the OP VVV project CZ.02.1.01/0.0/0.0/16019/000076 Research Center for Informatics.
}% <-this % stops a space
\thanks{$^{1}$Levente Hajder is with the Department of Algorithms and their Applications,
        Eotvos Lorand University, Budapest, Hungary
        {\tt\small hajder@inf.elte.hu}}%
        \thanks{$^{2}$Daniel Barath is with Visual Recognition Group, Department of Cybernetics, Czech Technical University in Prague, Czech Republic and with the
        Machine Perception Research Laboratory, MTA SZTAKI, Budapest, Hungary 
        {\tt\small barath.daniel@sztaki.mta.hu}}%
}

\begin{document}

\maketitle
\thispagestyle{empty}
\pagestyle{empty}

%%%%%%%%%%%%%%%%%%%%%%%%%%%%%%%%%%%%%%%%%%%%%%%%%%%%%%%%%%%%%%%%%%%%%%%%%%%%%%%%
\begin{abstract}
Two solvers are proposed for estimating the extrinsic camera parameters from a single affine correspondence assuming general planar motion. In this case, the camera movement is constrained to a plane and the image plane is orthogonal to the ground. The algorithms do not assume other constraints, e.g.\ the non-holonomic one, to hold.
A new minimal solver is proposed for the semi-calibrated case, i.e. the camera parameters are known except a common focal length. Another method is proposed for the fully calibrated case. 
Due to requiring a single correspondence, robust estimation, e.g. histogram voting, leads to a fast and accurate procedure. 
The proposed methods are tested in our synthetic environment and on publicly available real datasets consisting of videos through tens of kilometers. They are superior to the state-of-the-art both in terms of accuracy and processing time.
\end{abstract}

%%%%%%%%%%%%%%%%%%%%%%%%%%%%%%%%%%%%%%%%%%%%%%%%%%%%%%%%%%%%%%%%%%%%%%%%%%%%%%%%
\section{INTRODUCTION}

The estimation of the epipolar geometry of a stereo image pair is a fundamental problem for recovering the relative camera motion. Being a well-studied problem, several papers discussed its theory and applications. The epipolar geometry of a stereo image pair is described by a $3 \times 3$ fundamental matrix~\cite{hartley2003multiple}. 
In the case of known camera parameters, additional geometric constraints transform it to an essential matrix. 
In this paper, we focus on a special case when the optical axes of the cameras are in the same 3D plane which is parallel to the ground. 
Two algorithms are proposed exploiting a complex, and thus more informative, input, an affine correspondence (AC). The first one requires the cameras to be calibrated for estimating the essential matrix; whilst the second one solves the semi-calibrated case, i.e.\ the calibration is known except a common focal length.
In particular, we are interested in the minimal case, i.e.\ estimating the camera \textit{motion from a single correspondence}. 

%\textit{Literature overview.} 
An affine correspondence consists of a point pair and the related local affine transformation mapping the infinitely close vicinity of the point in the first image to the second one. Nowadays, several approaches are available for estimating ACs accurately. Besides the well-known affine-covariant feature detectors~\cite{mikolajczyk2005comparison} such as MSER~\cite{MSER}, Hessian-Affine, Harris-Affine~\cite{mikolajczyk2005comparison}, there are some recent approaches based on view-synthesizing, e.g.\ ASIFT~\cite{Morel2009}, ASURF, or MODS~\cite{mishkin2015mods}.
These methods obtain accurate local affinities by transforming the original image by affine transformations, creating synthetic views. Then a feature detector is applied to the warped images. 
The final local affinity is calculated as the combination of the two affine transformations originating from the view synthesis and the applied detector. Also, there is a recent method~\cite{mishkin2018repeatability} using deep network for recovering the affine frames.

Besides the widely-used point-based approaches for estimating epipolar geometry, e.g.\ the 5-~\cite{philip1996non,nister2004efficient,li2006five,batra2007alternative}, 6-~\cite{wang1997six,stewenius2008minimal,li2006simple,hartley2012efficient}, 7- and 8-point~\cite{hartley2003multiple} algorithms, there are recent ones exploiting ACs.  
Perdoch et al.~\cite{PerdochMC06} proposed two techniques for approximating the pose using 2 and 3 correspondences. 
Bentolila and Francos~\cite{Bentolila2014} proposed a method to estimate fundamental matrix $\Fund$ from 3 ACs using conic constraints. Raposo et al.~\cite{Raposo2016} suggested a solution for essential matrix estimation from 2 ACs. Bar\'ath et al.~\cite{barath2017minimal} showed that the relationship of affine correspondences and $\Fund$ is linear and geometrically interpretable. 
Using this linear relationship, they estimated the essential matrix and the common focal length from 2 ACs. 
These methods suppose that the camera movement is general.
Therefore, they do not exploit the additional constraints implied by special movement types, e.g.\ a camera mounted to a car.

Nowadays, as the attention is pointing towards autonomous driving, it is becoming more and more important to design algorithms which exploit the properties of such a movement and therefore provide results superior to the general solutions.
Considering that the cameras are moving on a plane, e.g.\ the camera is mounted to a car, is a well-known approach for reducing the degrees-of-freedom and thus speeding up the robust estimation. 
Ortin and Montiel~\cite{Ortin2001} showed that this problem can be solved exploiting two point correspondences. 
Since then, other solvers were proposed to estimate the motion from two correspondences~\cite{chou2015,choi2018}. 
Scaramuzza~\cite{Scaramuzza2011} proposed a technique using a single point pair for a special camera setting and he assumes the non-holonomic constraint.  
In brief, this constraint states that the camera can only rotate around an exterior point defined by the intersection lines drawn by the orientation of the wheels. Straight driving is interpreted by a point in the infinity, it is the movement on a circle with infinite radius. 
This constraint is valid only if the steering angle is constant in two images. It nevertheless holds only in ideal circumstances.  

\begin{figure}[h]
  	\centering
  	\includegraphics[width=0.70\columnwidth]{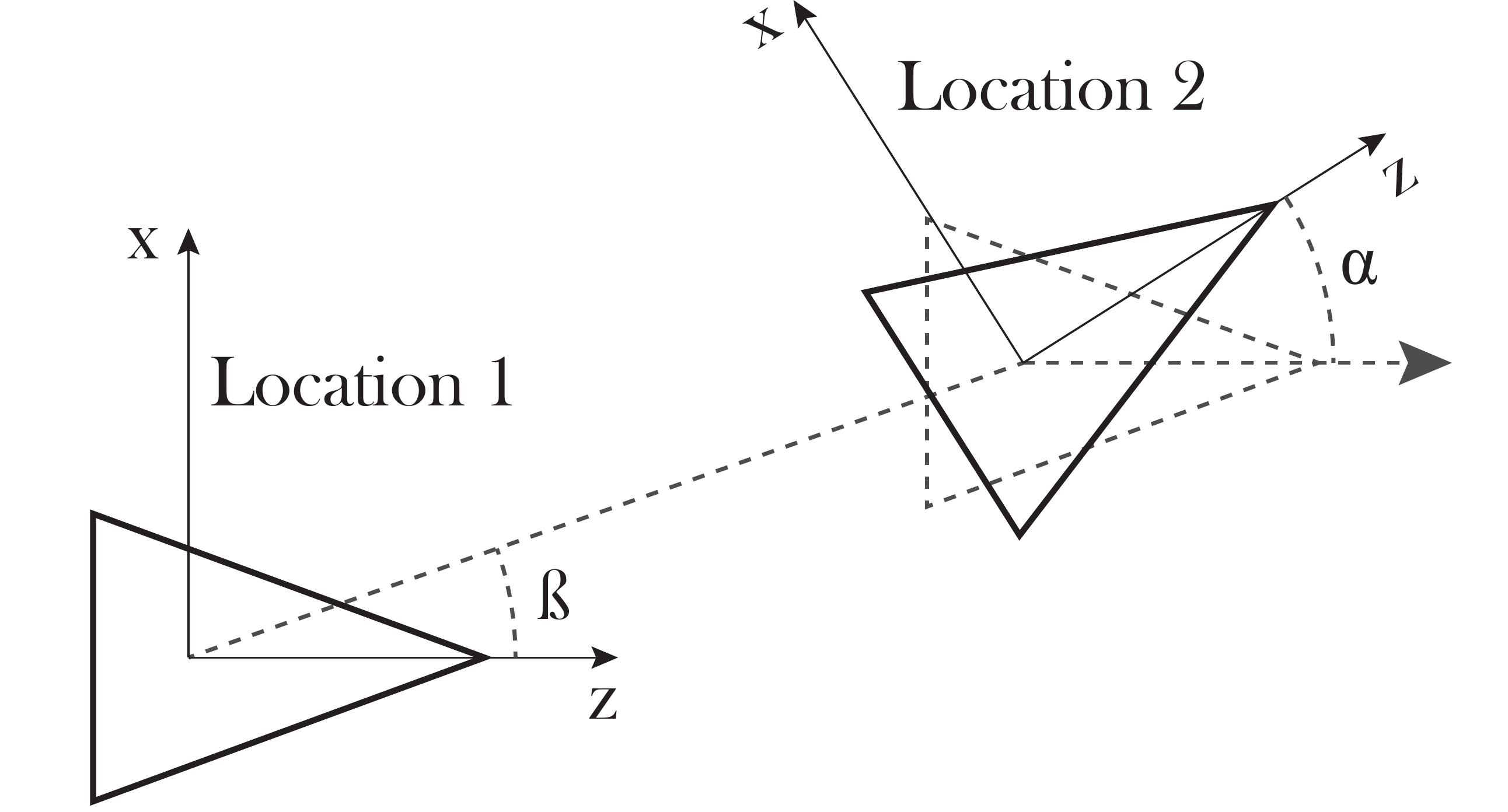}
  	\caption{Motion scheme. The camera movement is described by the angle $\alpha$ of the rotation perpendicular to axis $\text{Y}$ and translation vector $[\cos(\beta), \; 0 \;, \sin(\beta)]^\trans$. }
	\label{fig:illustration_image}
\end{figure}

\textit{Contributions.} 
An approach is proposed in this paper specializing the general relationship of affine correspondences and epipolar geometry~\cite{Raposo2016,barath2017minimal}. 
Since the cameras are supposed to move on a plane, the degrees-of-freedom are reduced to two:
the angle of the translation vector orthogonal to the vertical direction, and the rotation around the vertical axis, see Fig.~\ref{fig:illustration_image}. 
The proposed specialization makes the relative motion estimable from a single affine correspondence. 
To the best of our knowledge, this is the first minimal method that requires only a single correspondence to estimate the \textit{general planar motion}.
It is shown that even the semi-calibrated case, i.e.\ when the calibration is known except a common focal length, is solvable from a single AC. Applying the proposed one-point technique, robust estimators, e.g. histogram voting, is straightforwardly applicable.

\section{Notation and geometric background}
\label{sec:notations}

\noindent \textit{Affine Correspondences.}
In this paper, we consider an affine correspondence (AC) as a triplet: $(\Point_1, \Point_2, \Aff)$, where $\Point_1 = [p_{1x} \; p_{1y} \; 1]^\trans$ and $\Point_2 = [p_{2x} \; p_{2y} \; 1]^\trans$ are a corresponding homogeneous point pair in the two images, and 
$
	\Aff
$
is a $2 \times 2$ linear transformation which we call \textit{local affine transformation} with elements $a_1, a_2, a_3$, and $a_4$ (in row-major order). 
It transforms the neighboring pixels of $\Point_1$ in the first image to the pixels around $\Point_2$ in the second image. 
% patch points from the first to the second image if the coordinates are given in inhomogeneous form. 
Note that $\Aff$ is defined as the first-order Taylor-approximations of the $\text{3D} \to \text{2D}$ projection functions~\cite{barath2016novel}.
% Remark that, for perspective cameras, $\Aff$ is the first-order approximation of the related image-image transformation represented by a \textit{homography} if the tangent plane of the observed spatial object is considered.

\noindent \textit{Fundamental and essential matrices.}
The $3 \times 3$ fundamental matrix $\matr{F}$ is a projective transformation ensuring the epipolar constraint~\cite{hartley2003multiple} as $\matr{p_2}^\trans \matr{F} \matr{p_1} = \matr{p_2}^\trans \Intrinsic_2^{-\trans} \matr{E} \Intrinsic_1^{-1} \matr{p_1} = 0$. The relationship of essential matrix $\matr{E}$ and $\matr{F}$ is $\matr{F} = \Intrinsic_2^{-\trans} \matr{E} \Intrinsic_1^{-1}$, where $\Intrinsic_1$ and $\Intrinsic_2$ are the upper-triangular matrix consisting of the intrinsic parameters of the cameras.
In the rest of the paper, we assume points $\matr{p_1}$ and $\matr{p_2}$ to be premultiplied by  $\Intrinsic_1^{-1}$ and $\Intrinsic_2^{-1}$, simplifying the epipolar constraint to 
\begin{equation}
	\label{eq:epipolar_constraint}
	\matr{q_2}^\trans \matr{E} \matr{q_1} = 0,
\end{equation}
where $\matr{q}_1=\Intrinsic_1^{-1} \matr{p}_1= [q_{1x}\; q_{1y} \; 1]^\trans$ and $\matr{q_2} =\Intrinsic_2^{-1} \matr{p_2}= [q_{2x}\; q_{2y} \; 1]^\trans$ are the normalized point coordinates. Essential matrix $\matr{E}$ is described by the camera motion as follows: $\matr{E} = [\matr{t}]_\times \matr{R}$, where $\matr{t}$ is a 3D translation vector and $\matr{R}$ is an orthonormal rotation matrix. Operator $[.]_\times$ is the cross-product matrix. 
The $i$th element of the essential matrix in row-major order is denoted as $e_i$ ($i \in [1,9]$). 

\noindent
\textit{Planar motion.}
We are given a calibrated image pair with a common plane $Y=0$, if axis $\text{Y}$ is parallel to the vertical direction of the image planes. Thus, the vertical directions of the images are parallel. A trivial example for that constraint is the camera setting of autonomous cars: a camera is fixed to the car and the axis $\text{Y}$ of the cameras is perpendicular to the ground plane.
To estimate the camera motion, we first describe the parameterization of the motion.

Let us denote the 1st and 2nd projection matrices
by $\Proj_1$ and $\Proj_2$. Without loss of generality, the world coordinate system is fixed to the 1st camera. Thus, 
$
\Proj_1 = \Intrinsic_1 \; [ \; \matr{I}_{3 \times 3} \; | \; \textbf{0} \; ],
$
where $\Intrinsic_1$ is the intrinsic camera parameters of the 1st camera.
The 2nd one is
$
\Proj_2 = \Intrinsic_2 \; [ \; \matr{R}_{2} \; | \; \matr{t}_{2} \; ],
$
where $\Intrinsic_2$, $\matr{R}_{2}$, and $\matr{t}_{2}$ are the intrinsic camera matrix, orientation and location of the second camera, respectively.
Assuming planar motion, the rotation and translation are represented by three parameters: a 2D translation and a rotation. Formally,
\[
\begin{array}{ccc}
\matr{R}_{2} = \left[\begin{array}{ccc}
 \cos  \alpha & 0 & \sin  \alpha \\
 0 & 1 & 0 \\
 - \sin  \alpha & 0 &  \cos  \alpha \\
\end{array}\right],
&~&
\matr{t} = \left[\begin{array}{c}
x\\
0\\
z
\end{array}\right],
\end{array}
\]
where $\alpha \in [0, 2 \pi)$ is the rotation around axis $\text{Y}$. Therefore, the essential matrix~\cite{hartley2003multiple} $\Ess = \left[ \matr{t} \right]_{\times} \matr{R}_{2}$ is simplified as follows: 
\begin{align}
    \begin{split}
        \Ess =  
        \left[\begin{array}{ccc}
        	0 & -z & 0 \\ 
        	z \cos \alpha + x \sin \alpha & 0 & z \sin \alpha -x \cos \alpha\\
        	0 & x & 0
    	\end{array}\right]
    \end{split}.
    \label{eq:E}
\end{align}
Thus, $e_{1} = e_{3} = e_{5} = e_{7} = e_{9} = 0$, $e_{2} = -z$, $e_{8} = x$,
$e_{4} = z \cos \alpha - x \sin  \alpha $, and $e_{6} = -z \sin  \alpha - x  \cos  \alpha$. As it is well known in projective geometry~\cite{hartley2003multiple}, the scale of the motion cannot be recovered. Therefore the planar translation parameters $x$ and $z$ are described by the coordinates of a point on a unit-circle as follows: $x = \cos \beta$ and $z = \sin \beta$, $\beta \in [0, 2 \pi)$. The non-zero elements of $\Ess$ are rewritten as
\begin{equation*}
	\begin{array}{lcl} 
	e_6 = \sin \beta \sin  \alpha - \cos \beta  \cos  \alpha = - \cos \left( \alpha + \beta \right),\\ 
	e_4 =  \sin \beta \cos \alpha + \cos \beta \sin  \alpha = \sin \left( \alpha + \beta \right),\\
    e_8 =  \cos \beta, \quad
	e_2 = - \sin \beta.
  	\end{array}
%	\label{eq:E_planar_elements}
\end{equation*}
Consequently, the motion has two degrees of freedom (DoF): the angles of the rotation and translation.  

\section{Pose from an Affine Correspondence}
\label{sec:proposed_method}
 
In this section, the proposed AC-based motion estimation is shown. The input is an affine correspondence, i.e.\ a point pair $\Qoint_1$, $\Qoint_2$ and the related local affine transformation $\Aff$. 

\subsection{Constraints from a single correspondence}
\label{sec:all_const}
 
\noindent
\textbf{For a point correspondence} consistent with $\Ess$, the following \textit{epipolar constraint}~\cite{hartley2003multiple} holds: $\Qoint_{2}^\trans \Ess \Qoint_{1} = 0$. This can be written as follows:
\begin{equation}
  \label{eq:sol_pts}
  	\begin{bmatrix}  -q_{2y} &  q_{1x} q_{2y} &   q_{1y} & -q_{2x} q_{1y} \end{bmatrix} \matr x = 0,
\end{equation}
where $\matr x = \begin{bmatrix}  \cos \left( \alpha + \beta \right) & \sin \left( \alpha + \beta \right) & \cos \beta  &  \sin \beta  \end{bmatrix}^\trans$.

\noindent
\textbf{The constraints from an affine transformation} 
implied on $\Ess$ are written by two linear equations \cite{barath2017minimal}. 
In the case of the discussed planar motion, several elements are zero in $\Ess$ and, thus, the linear system of \cite{Raposo2016} is simplified. The equations are as follows: 
\begin{align}
    \begin{split}
        a_{3} e_{6} + \left(q_{2y} + a_{3} q_{1x} \right) e_{4}  + a_{1} q_{1y} e_{2}= 0, \\
        a_{4} e_{6} + a_{4} q_{1x} e_{4} + e_{8} + \left(q_{2x} + a_{2} q_{1y} \right) e_{2} = 0.
    \end{split}
    \label{eq:sol_aff}
\end{align}

\noindent
\textbf{From an affine correspondence}, the three constraints of Eqs. \ref{eq:sol_pts}, \ref{eq:sol_aff} are combined as $\matr B \matr x = 0$, where 
\begin{eqnarray}
    \matr B = 
    \begin{bmatrix} 
           -q_{2y} &   q_{1x} q_{2y} &  q_{1y} & -q_{2x} q_{1y}\\
         -a_3 & q_{2y} + a_3 q_{1x}  &  0 & -a_{1} q_{1y} \\
         -a_4 & a_{4} q_{1x} &  1 & -q_{2x} - a_{2} q_{1y}
        \end{bmatrix}
    \label{eq:sol_total}
\end{eqnarray}
is the coefficient matrix and $\matr x = \begin{bmatrix} -e_6 & e_4 & e_8 & -e_2 \end{bmatrix}^\trans = \begin{bmatrix}   \cos \left( \alpha + \beta \right) & \sin \left( \alpha + \beta \right) & \cos \beta & \sin \beta  \end{bmatrix}^\trans$ is the vector of unknowns. 
In this case, $\matr x$ is the null-vector of $\matr B$.

\subsection{Calibrated case}
\label{sec:cal_case}

If the cameras are calibrated, the problem is to estimate the two unknown angles, $\alpha$ and $\beta$, from the three constraints described previously. 
For this case, several solvers exist considering planar motion~\cite{chou2015,choi2018}. 
We chose the solver called "Line" of Choi et al.~\cite{choi2018}. 
The proposed method that estimates the essential matrix from an affine correspondence is called 1AC in the rest of the paper.

The null space of Eq.~\ref{eq:sol_total} can be determined if at least three correspondences are given. In this case, $\matr B$ has three rows, and each of them originates from a correspondence via Eq.~\ref{eq:sol_pts}. 
This linear three-point algorithm is called 3PC in the experiments.

\subsection{Semi-calibrated case}
\label{sec:focal_solver}

When semi-calibrated cameras are given, the intrinsic camera matrices are $\matr{C}_1 = \matr{C}_2 = \text{diag}(f,f,1)$, 
where $f$ is the unknown focal length. Fundamental matrix $\Fund$ is 
\[
\small
\Fund = \matr{C}_2^{-\trans} \Ess \matr{C}_1^{-1} = \left[\begin{array}{ccc}
 0 & - \frac{\sin \beta}{f^2}  & 0 \\
 \frac{\sin \left( \alpha + \beta \right)}{f^2} & 0 & - \frac{\cos \left( \alpha + \beta \right)}{f} \\
0 & \frac{\cos \beta}{f} & 0 \\
 \end{array}\right].
\]
Considering that there is an unknown scale $\mu$, an equation system is formed from the elements of the fundamental matrix and the null-vector of Eq.~\ref{eq:sol_total}\footnote{Vector $\matr x = \begin{bmatrix} -f_6 & f_4 & f_8 & -f_2 \end{bmatrix}^\trans = \begin{bmatrix}   \cos \left( \alpha + \beta \right)/f & \sin \left( \alpha + \beta \right)/f^2 & \cos \beta /f & \sin \beta /f^2  \end{bmatrix}^\trans$ for the semi-calibrated case.} as follows: 
\begin{equation}
    \small
    \label{eq:nullvector}
    \frac{1}{\mu} \left[\begin{array}{c}
        f^{-1}\cos(\alpha + \beta)\\
        f^{-2}\sin(\alpha + \beta)\\
        f^{-1}\cos\beta\\
        f^{-2}\sin\beta
    \end{array}\right] = \left[\begin{array}{c}
        n_{1}\\
        n_{2}\\
        n_{3}\\
        n_{4}
    \end{array}\right],
\end{equation}
where $[n_1 \; n_2 \; n_3 \; n_4]^\trans$ is the null-vector. 
Dividing the first equation in Eq.~\ref{eq:nullvector} by the third one leads to the following formula: 
$\frac{\cos(\alpha + \beta)}{\cos\beta} = \frac{n_{1}}{n_{3}}$. 
After rearranging, it becomes $n_{3}\cos(\alpha + \beta) = n_{1}\cos\beta$. 
Due to having trigonometric functions, it is easier to solve the system if the equation is squared. It is as follows:
$
    n_{3}^{2}\cos^{2}(\alpha + \beta)=n_{1}^{2}\cos^{2}\beta.
$
Similarly, after dividing the second equation by the fourth one, getting its square, and replacing $\sin^2(.)$ by $1 - \cos^2(.)$, the following equation is given: 
$
    n_{4}^{2}\left(1-\cos^{2}(\alpha + \beta)\right)=n_{2}^{2}\left(1-\cos^{2}\beta\right).
$
These equations can be written in matrix form as follows:
\begin{equation*}
\left[\begin{array}{cc}
n_{3}^{2} & -n_{1}^{2}\\
-n_{4}^{2} & n_{2}^{2}
\end{array}\right]\left[\begin{array}{c}
\cos^{2}(\alpha + \beta)\\
\cos^{2}\beta
\end{array}\right]=\left[\begin{array}{c}
0\\
n_{2}^{2}-n_{4}^{2}
\end{array}\right].
\end{equation*}

If the squares of angles $\beta$ and $\left( \alpha + \beta \right)$ are obtained, the four candidate values for the angles are calculated as $\beta_1=-\beta_2=\cos^{-1}(\sqrt{(\cos^2 \beta)})$, and $\beta_3=-\beta_4=\cos^{-1}(-\sqrt{(\cos^2\beta)})$. 
There are four candidates also for angle $\alpha + \beta$. 
Therefore, in total, $16$ permutations of the two angles have to be considered. 
The focal length is obtained by dividing the 1st coordinate in Eq.~\ref{eq:nullvector} by the 2nd one after the angles are substituted. The division of the 3rd row by the 4th one also yields the focal length, the difference between the candidate focal length can be used to check the accuracy of the estimation. The focal length must be positive; this fact is also applied to reject invalid candidate values.

For noisy data, it may happen that the values obtained for $\cos^{2}(\alpha + \beta)$ and $\cos^{2} \beta$ are smaller than 0 or greater than 1.  
In this case, the closest valid solution is selected: $\pm \pi/2$ or $0$. 
 
\subsection{Outlier removal}

To apply the proposed one-point algorithm to real-world problems where the data might contain a significant number of outliers, the method has to be combined with a robust estimator. 
Benefiting from the fact that even \textit{a single correspondence is enough for the estimation}, weighted histogram voting~\cite{bujnak2009robust} is a justifiable choice. 
Using histogram voting has several beneficial properties compared to randomized algorithms, e.g.\ RANSAC~\cite{fischler1981random}. 
Most importantly, it is often superior both in terms of accuracy and processing time~\cite{bujnak2009robust,Scaramuzza2011}. 
The increased accuracy originates from the fact that it combines the information of all motion candidates and, thus, the outcome does not depend on a single sample where the model attained the maximum quality. 

Another advantageous aspect of applying histogram voting is the processing time depending linearly on the point number. 
The processing time of randomized algorithms, e.g.\ RANSAC, is unpredictable in most of the cases depending on both the number of points and the proportion of the outliers. 
For histogram voting, it depends only on the number of points and, therefore, the number of required model estimations is known even before the procedure is applied. Also, histogram voting can be easily parallelized both on multiple CPUs and GPU threads. 

%For the proposed voting algorithm, a histogram is generated consisting of $360$ bins, one for each integer degree. 
%Then the motion candidates are estimated separately from all of the affine correspondences. Each candidate increases the appropriate histogram bin. 
%Finally, the most dominant angle is selected as the highest peak. To overcome the discrete nature of histogram voting, the final solution is the weighted average of degrees around the selected peak with window size $s$ (we used $s = 2$ in the experiments). 

\begin{table*} 
	\center
	\resizebox{0.99\textwidth}{!}{\begin{tabular}{ c | c c c c | c c c c | c c c c | c c c c}
    \hline
   		\cellcolor{black!10} ~ & \multicolumn{4}{ c | }{ \cellcolor{black!10}\textbf{1ACf} } & \multicolumn{4}{ c |}{ \cellcolor{black!10}2PC~\cite{choi2018} and \textbf{1AC} } & \multicolumn{4}{ c | }{ \cellcolor{black!10}3PC~\cite{Bentolila2014} }  & \multicolumn{4}{ c }{ \cellcolor{black!10}5PC~\cite{nister2004efficient} } \\
    \hline
   		steps & \multicolumn{4}{ c | }{$4 \times 4$ EIG } & \multicolumn{4}{ c | }{$2 \times 2$ INV + $2 \times 2$ SVD.} & \multicolumn{4}{ c | }{$3$th degree poly. + conic intersect} & \multicolumn{4}{ c }{$5 \times 9$ SVD + $10 \times 10$ Gauss-J + $10 \times 10$ EIG} \\ 
   		1 iter & \multicolumn{4}{ c | }{$4^3  = 64$} & \multicolumn{4}{ c | }{$2^3 + 2^3 =16$} & \multicolumn{4}{ c | }{$3^2 \log(3) +12 \approx 22$} & \multicolumn{4}{ c }{$9 \cdot 5^2 + 10^3 + 10^3=2,225$ }\\ 
    \hline
   		$m$ & \multicolumn{4}{ c | }{1} & \multicolumn{4}{ c | }{2} & \multicolumn{4}{ c | }{3} & \multicolumn{4}{ c }{5}\\
    \hline
   	1 - $\mu$ & 0.25 & 0.50 & 0.75 & 0.90 & 0.25 & 0.50 & 0.75 & 0.90 & 0.25 & 0.50 & 0.75 & 0.90 & 0.25 & 0.50 & 0.75 & 0.90 \\ 
    \hline 
       	\# iters & 4 & 7 & 17 & 44 & 6 & 17 & 72 & 459 & 9 & 35 & 293 & 4,603 & 17 & 146 & 4,714 & $\sim 5 \cdot 10^5$  \\ 
       	\# comps & 256 & 448 & 1,088 & 2,816 & 96 & 272 & 1,152 & 7,344 & 198 & 770 & 6,446 & 101,266 & 37,825 & 324,850 & $\sim 10^7$  & $\sim 10^9$ \\ 
    \hline      
    \end{tabular}}
    \caption{The theoretical computational complexity of four different solvers (four blocks, each consisting of four columns) used in a RANSAC-like robust framework. The reported properties are: the number of operations of each solver (steps; $1$st row); the computational complexity of one estimation (1 iter; $2$nd); the number of correspondences required for the estimation ($m$; $3$rd); possible outlier ratios ($1 - \mu$; $4$th); the number of iterations needed for RANSAC with the required confidence set to $0.99$ (\# iters; $5$th); and computation complexity of the full procedure (\# comps; $6$th).}
\label{tab:computational_complexity}
\end{table*}

% \subsection{Remark on deep learning based odometry.} 
% This paper solves visual odometry by the application of projective geometry. Until now, two PCs have been required for this task, and the accuracy of the PC-based methods is in inverse ratio to the distances between the PCs. Recently, there are very efficient deep learning solutions~\cite{Wang2017,Zhan2018} for odometry. The operation of these networks for this task is not trivial as neural systems can connect the \textit{neighbouring pixels}, and accurate odometry requires \textit{far correspondences}. We think that the application of ACs explains this contradiction: a single affine correspondence determines the two parameters of a planar motion plus the focal length of the camera. We think that deep learning methods estimates the odometry parameters via affine transformations in their 'deep layers', and the application of neural networks for uncalibrated camera images is possible because the focal length can also be computed from an affinity.

\section{Experimental results}

Here, the proposed methods are tested in a synthetic environment and on publicly available real-world datasets.

\begin{figure}[h]
    \centering
	\includegraphics[width=0.40\columnwidth]{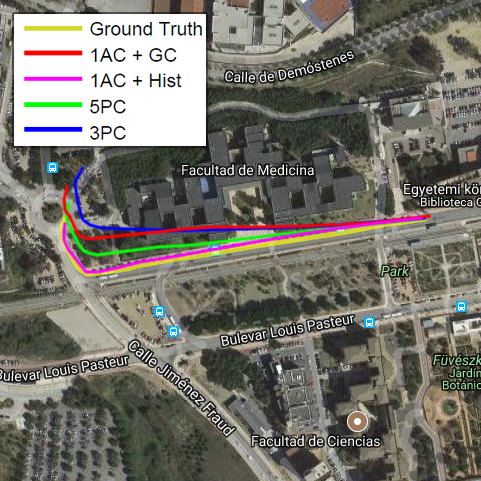}
	\includegraphics[width=0.40\columnwidth]{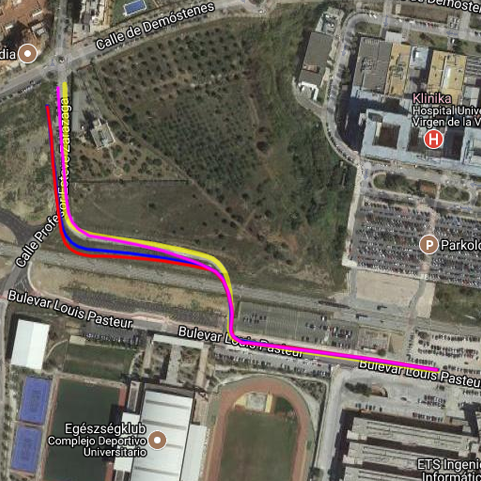}
\caption{Example paths from the Malaga dataset. Ground truth paths (yellow), proposed one-point algorithm with GC-RANSAC (red) and histogram voting (purple), three- (blue) and five-point (green) algorithms are visualized. }
\label{fig:example_results}
\end{figure}

\begin{figure}[h]
    \centering
	\includegraphics[width=0.85\columnwidth]{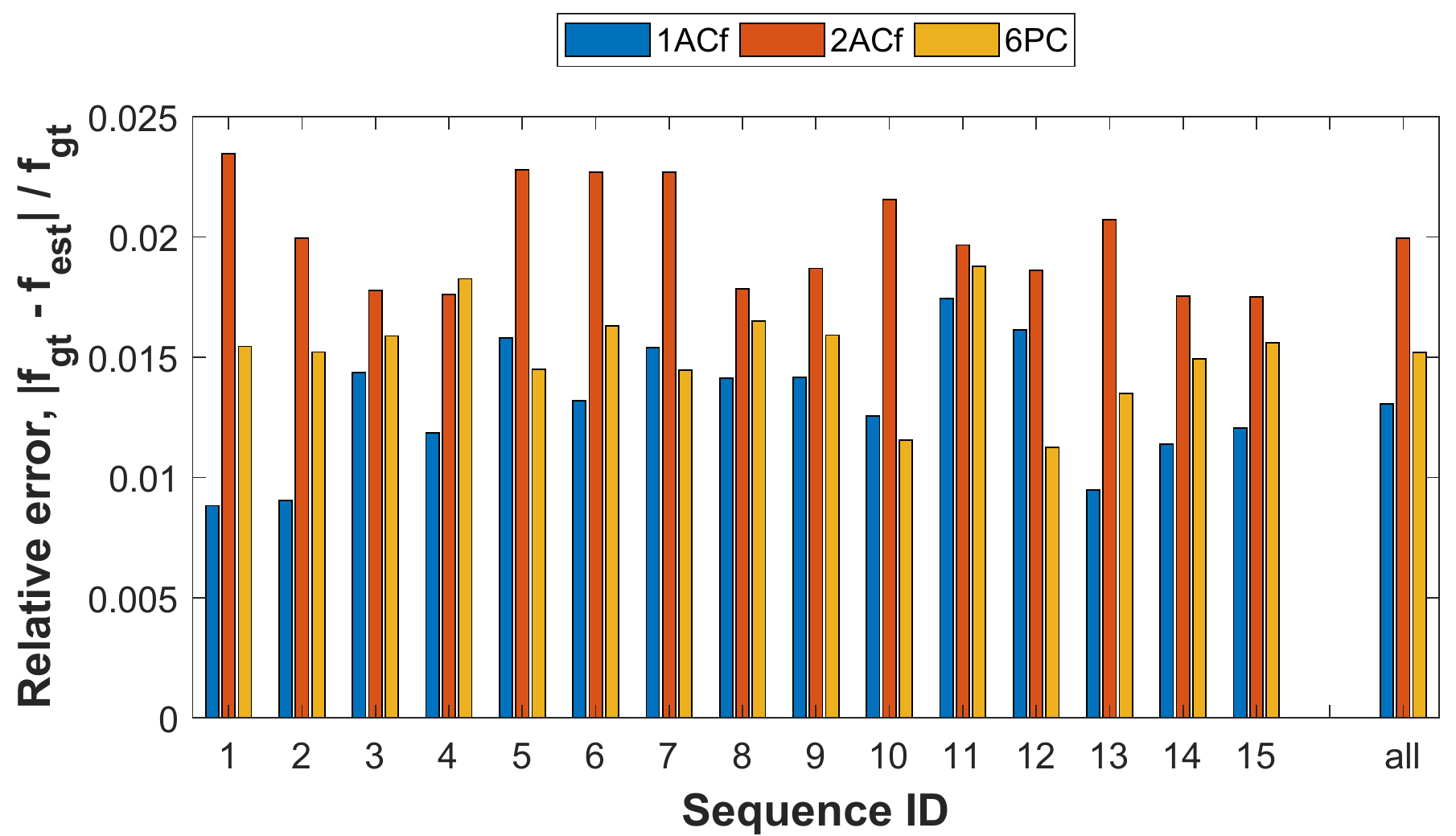}
    \caption{ The average relative focal length error, i.e.\ $|f_{\text{est}} - f_{\text{gt}}| / f_{\text{gt}}$, of the proposed (1ACf), the 2ACf~\cite{barath2017minimal} and 6PC~\cite{hartley2012efficient} methods on the $15$ sequences (horizontal; $6,111$ image pairs) of the Malaga dataset. The last triplet of bars reports the average error on all sequences. }
\label{fig:real_focal_errors}
\end{figure}

\subsection{Computational complexity}

\begin{figure*}[htb] 
	\centering
	\begin{subfigure}[t]{1.90\columnwidth}
        	\includegraphics[width = 0.245\columnwidth]{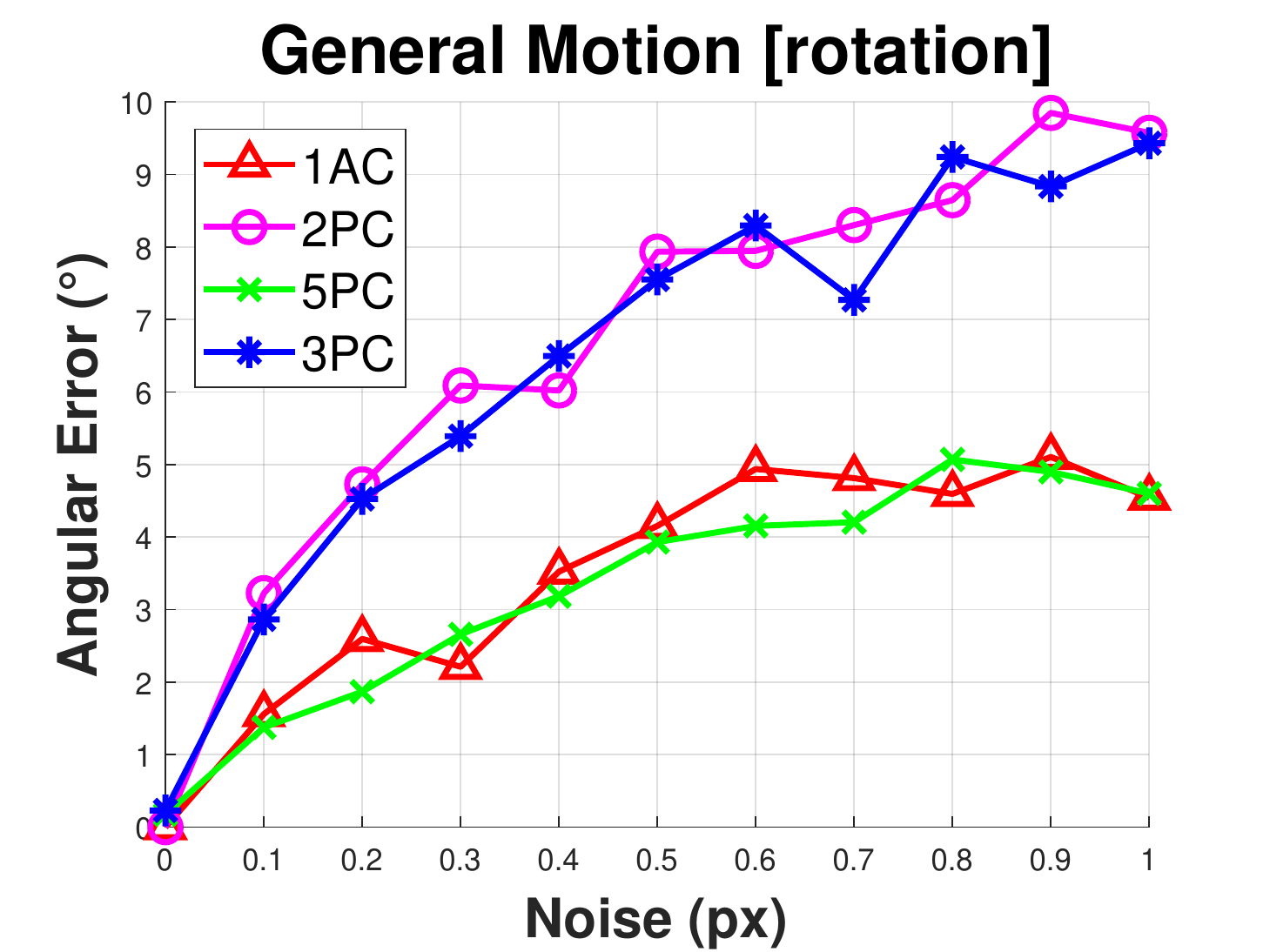}
        	\includegraphics[width = 0.245\columnwidth]{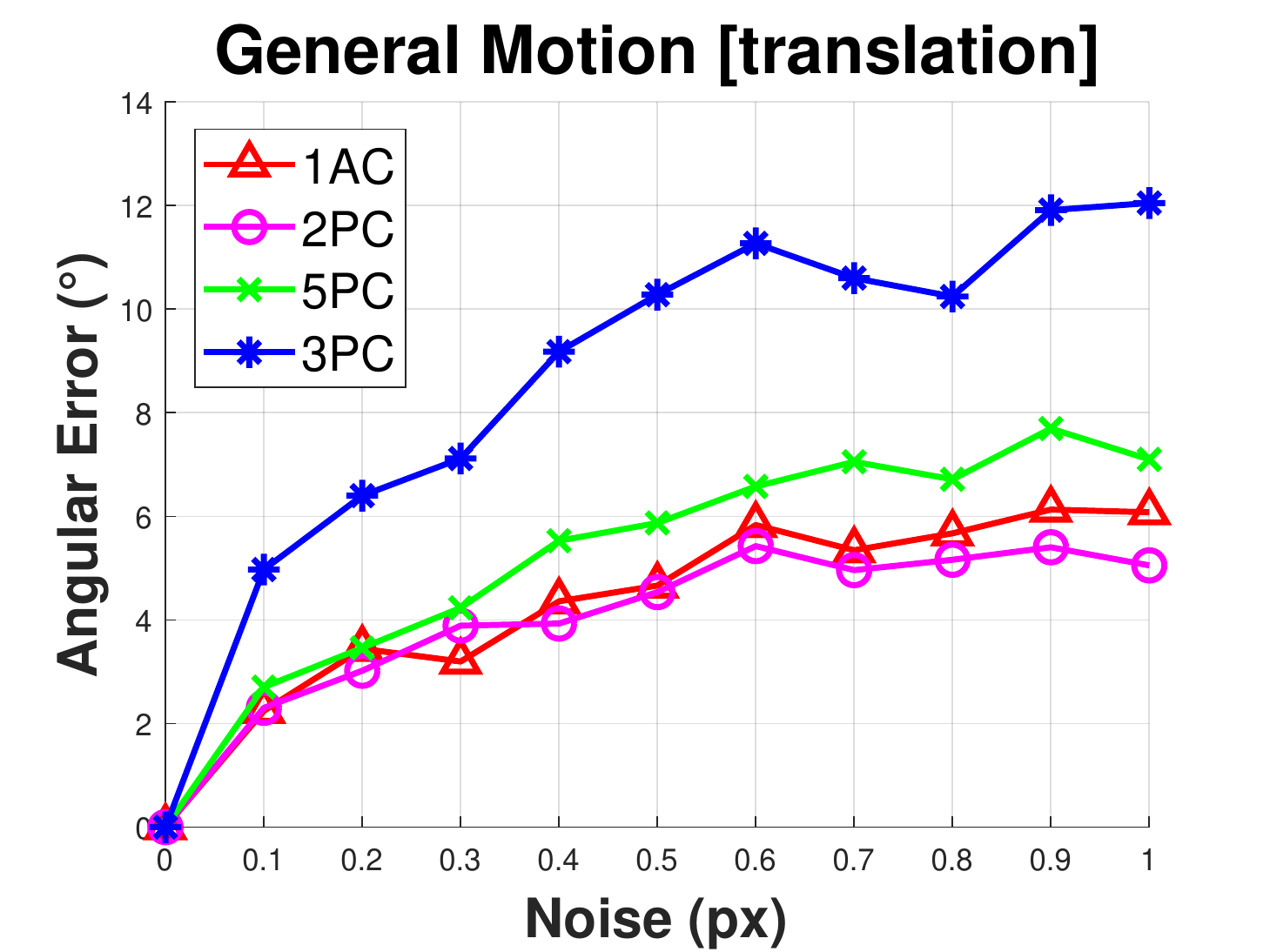} 
        	\includegraphics[width = 0.245\columnwidth]{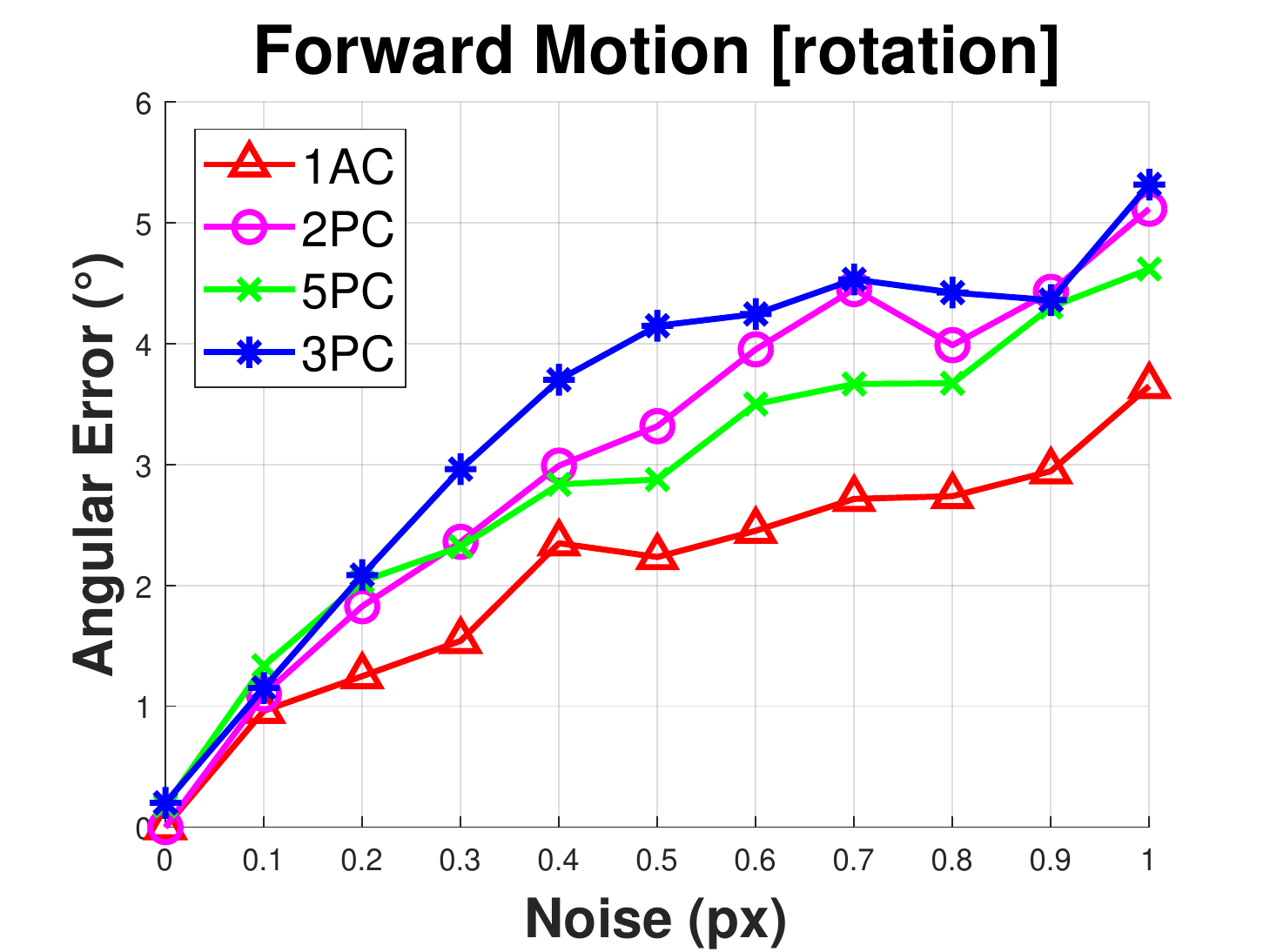}
        	\includegraphics[width = 0.245\columnwidth]{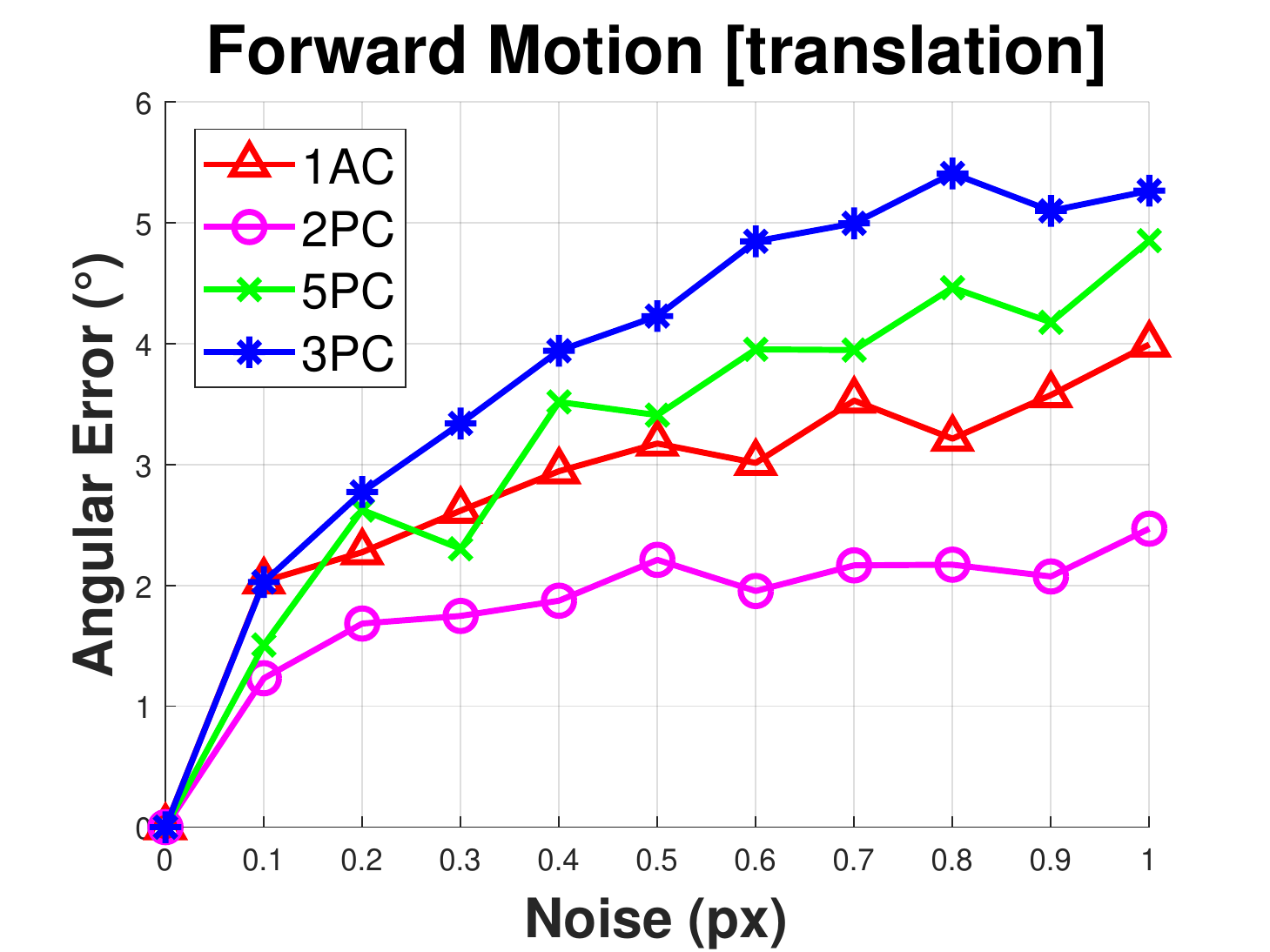} 
    		\caption{Calibrated cameras}
            \label{fig:synthesized_tests_calibrated}
	\end{subfigure}
	\begin{subfigure}[t]{1.90\columnwidth}
        	\includegraphics[width = 0.245\columnwidth]{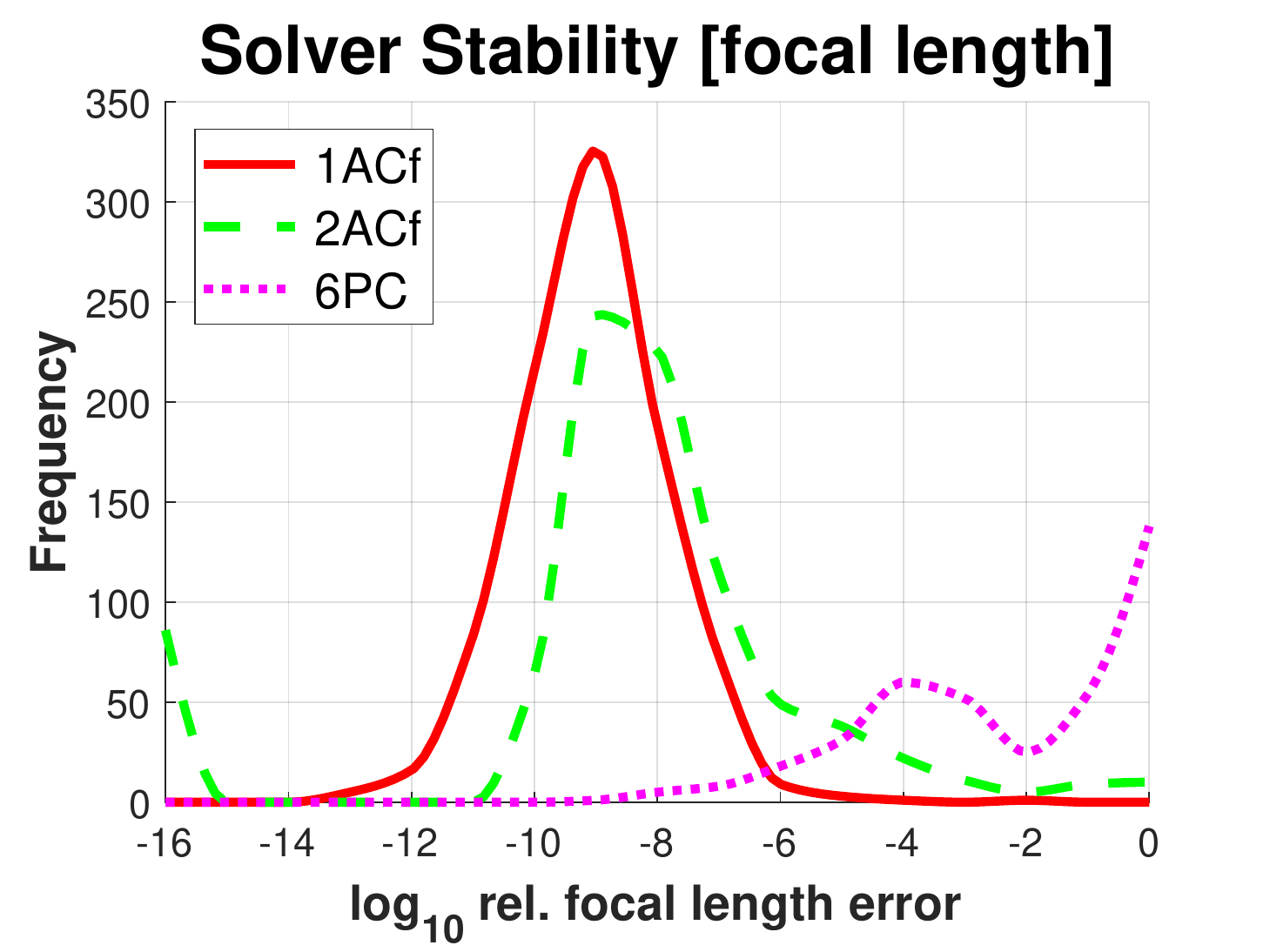}
        	\includegraphics[width = 0.245\columnwidth]{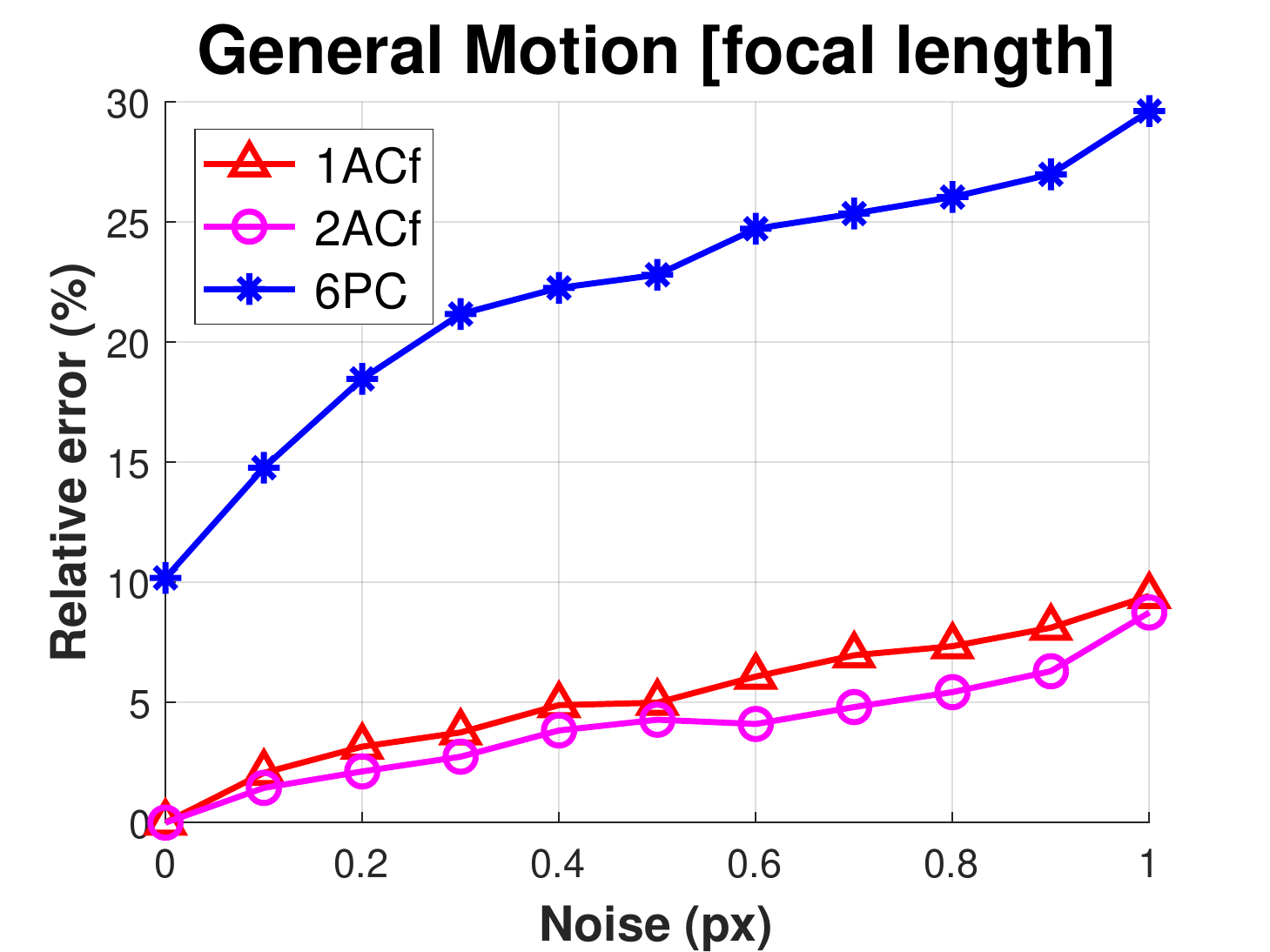} 
        	\includegraphics[width = 0.245\columnwidth]{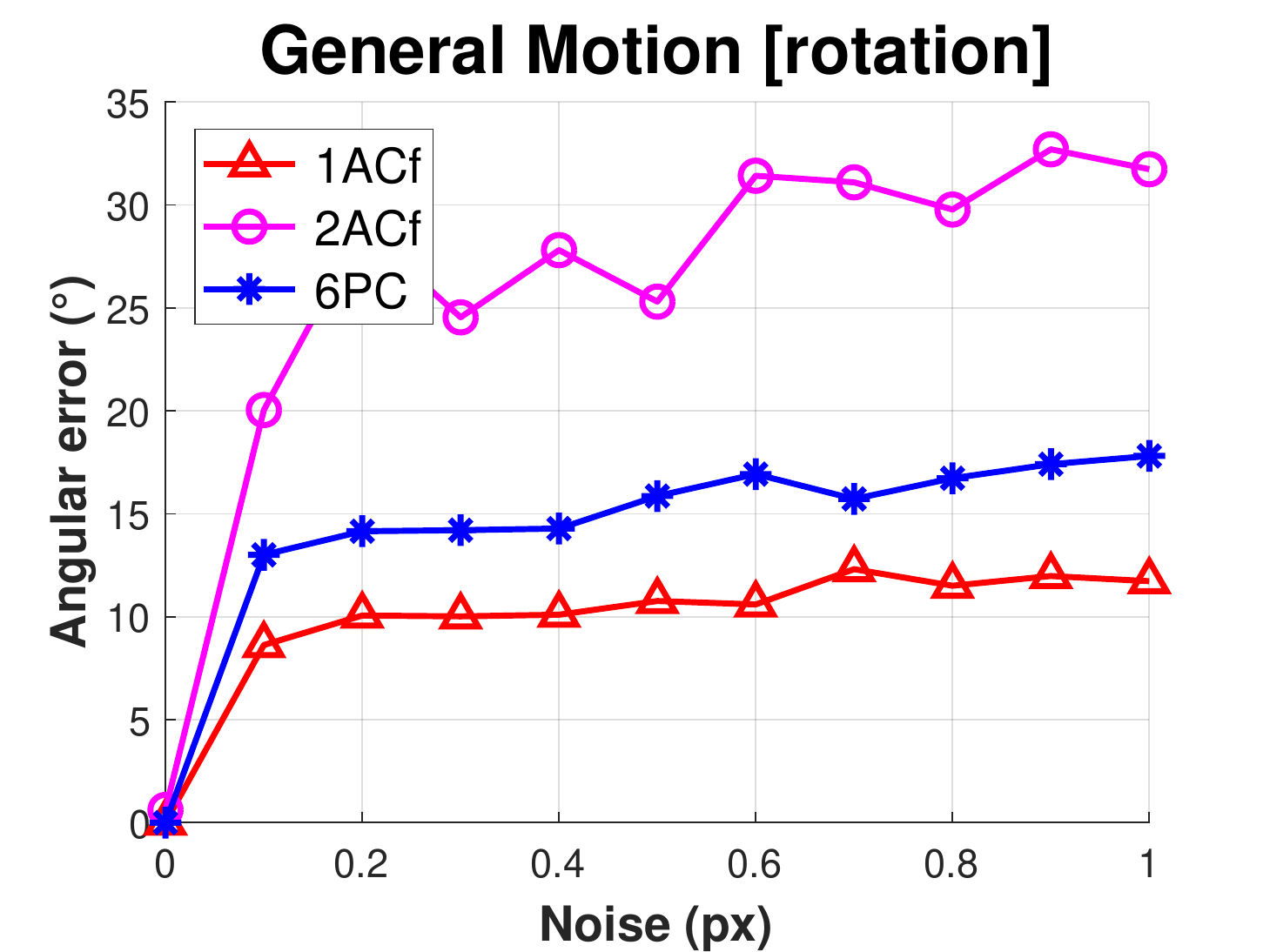}
        	\includegraphics[width = 0.245\columnwidth]{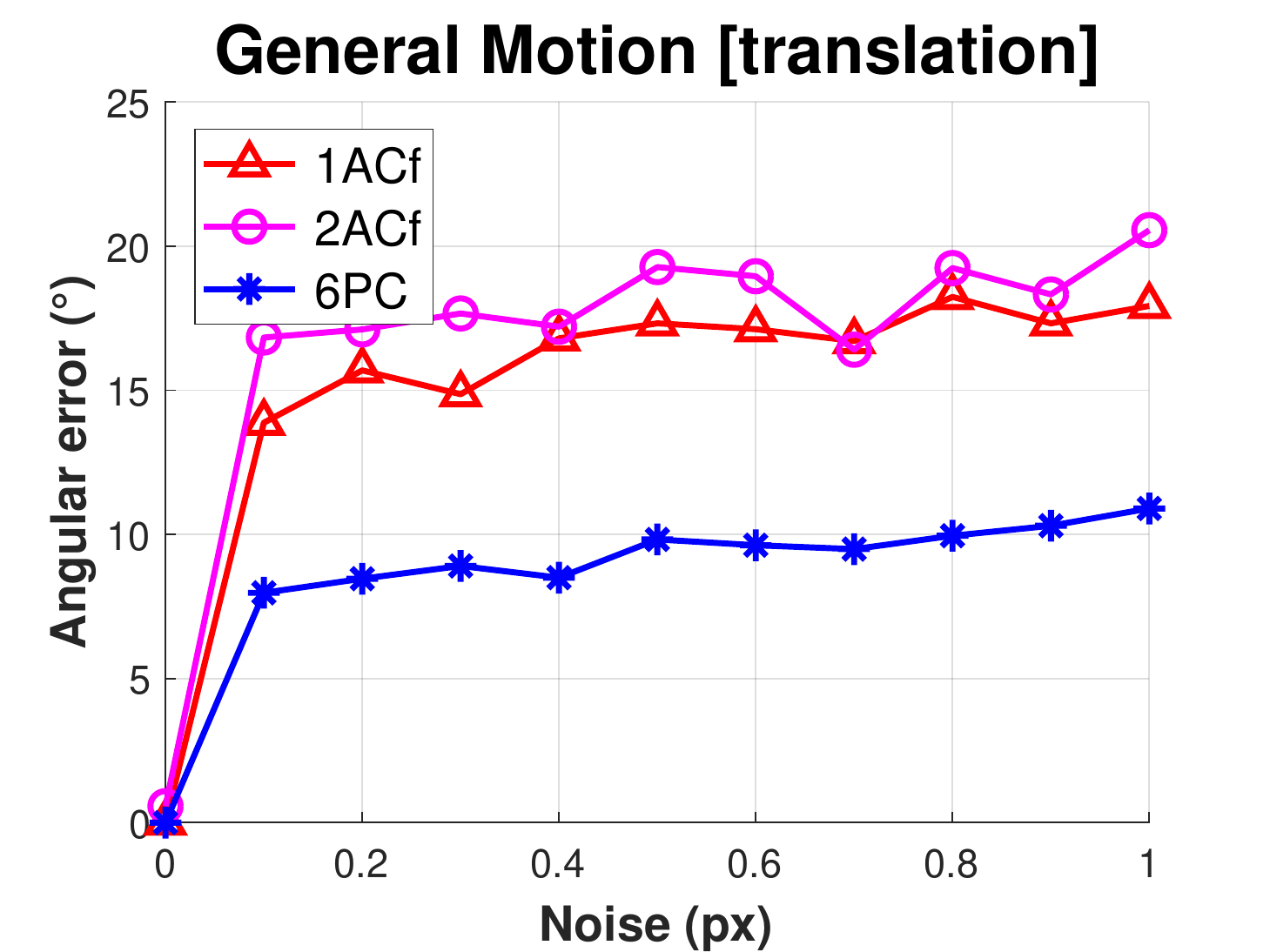} 
    		\caption{Unknown focal length}
            \label{fig:synthesized_tests_focal}
	\end{subfigure}
	\begin{subfigure}[t]{1.90\columnwidth}
        	\includegraphics[width=0.245\columnwidth]{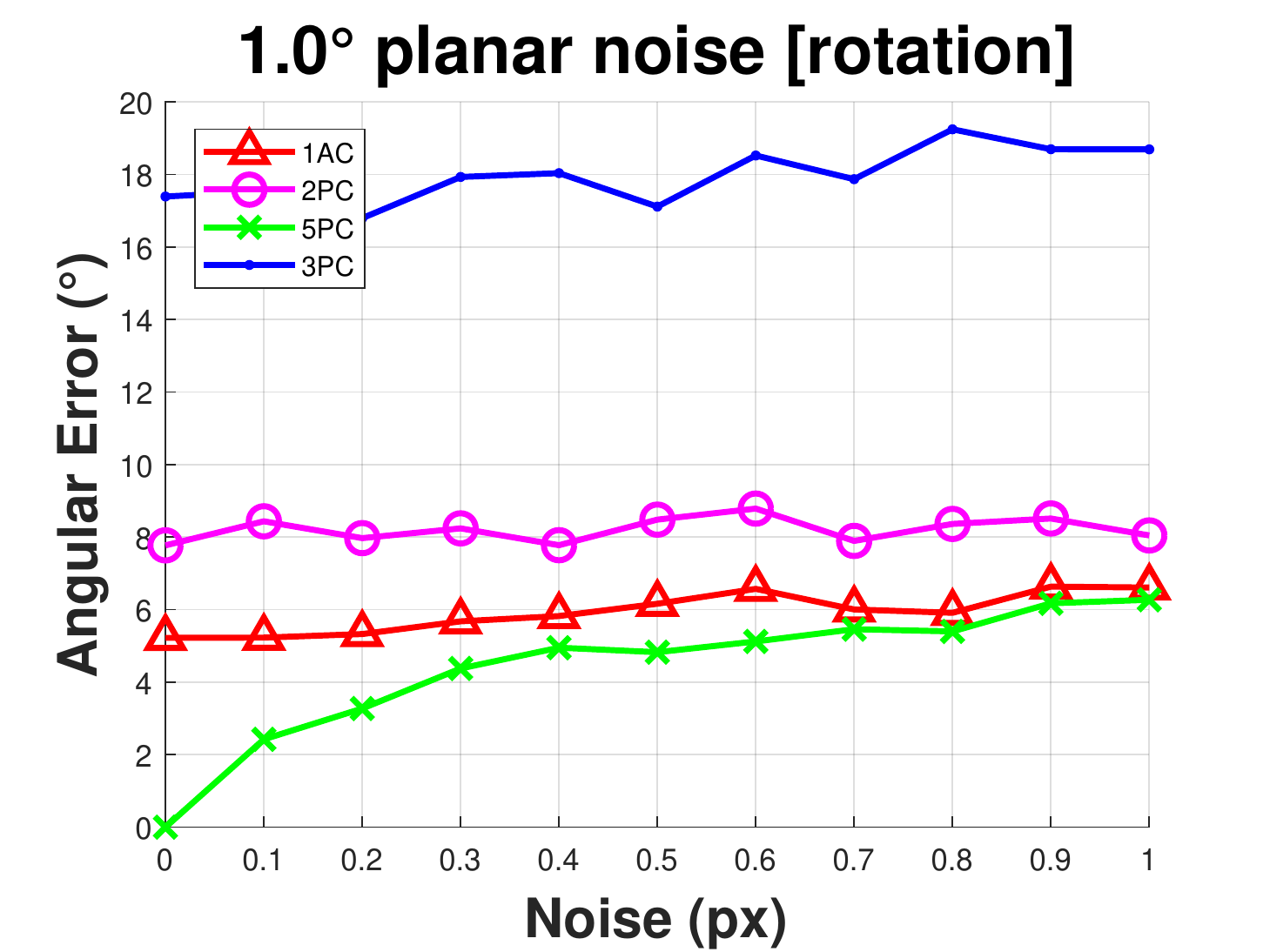}
        	\includegraphics[width=0.245\columnwidth]{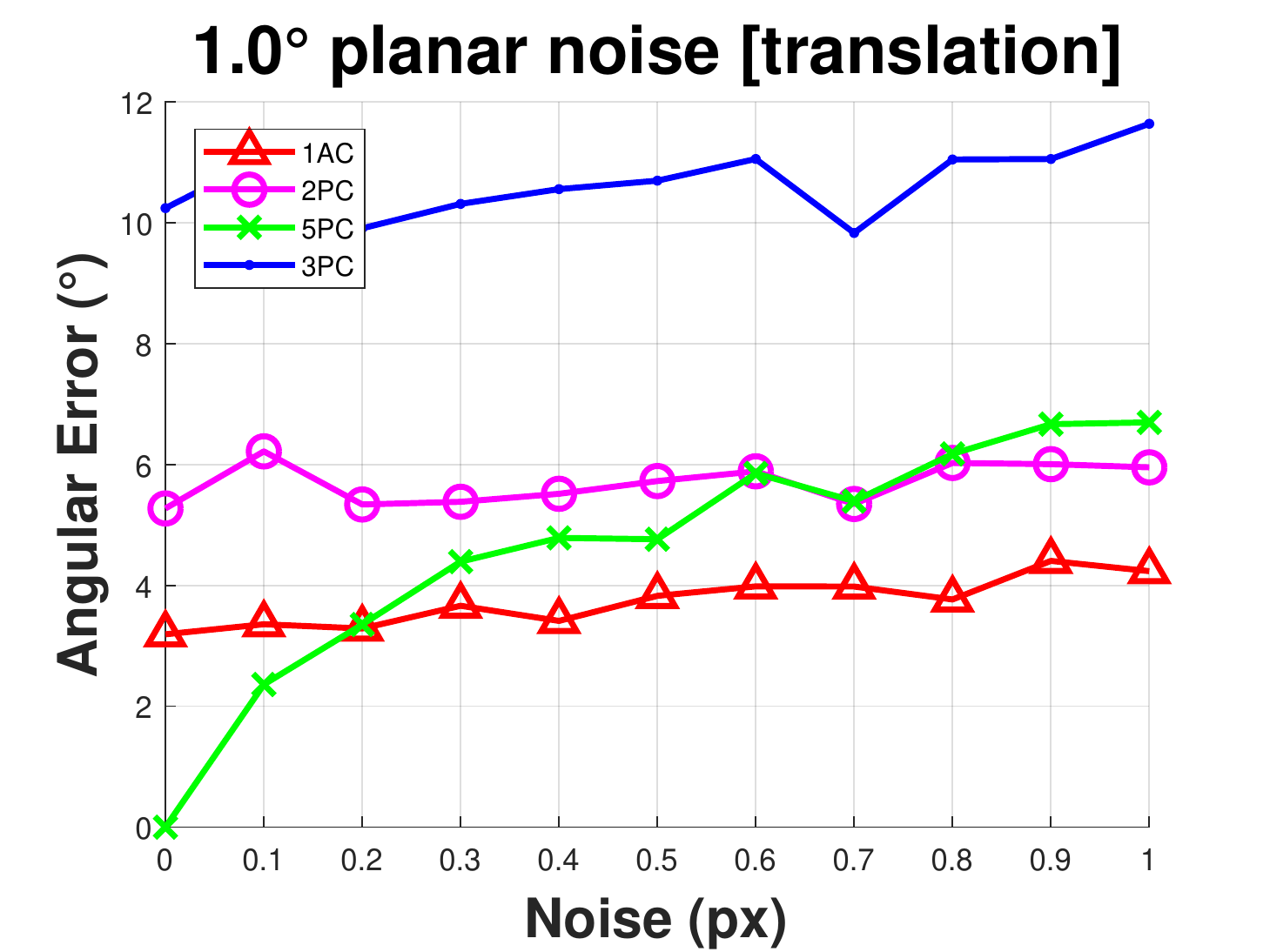}
            \includegraphics[width=0.245\columnwidth]{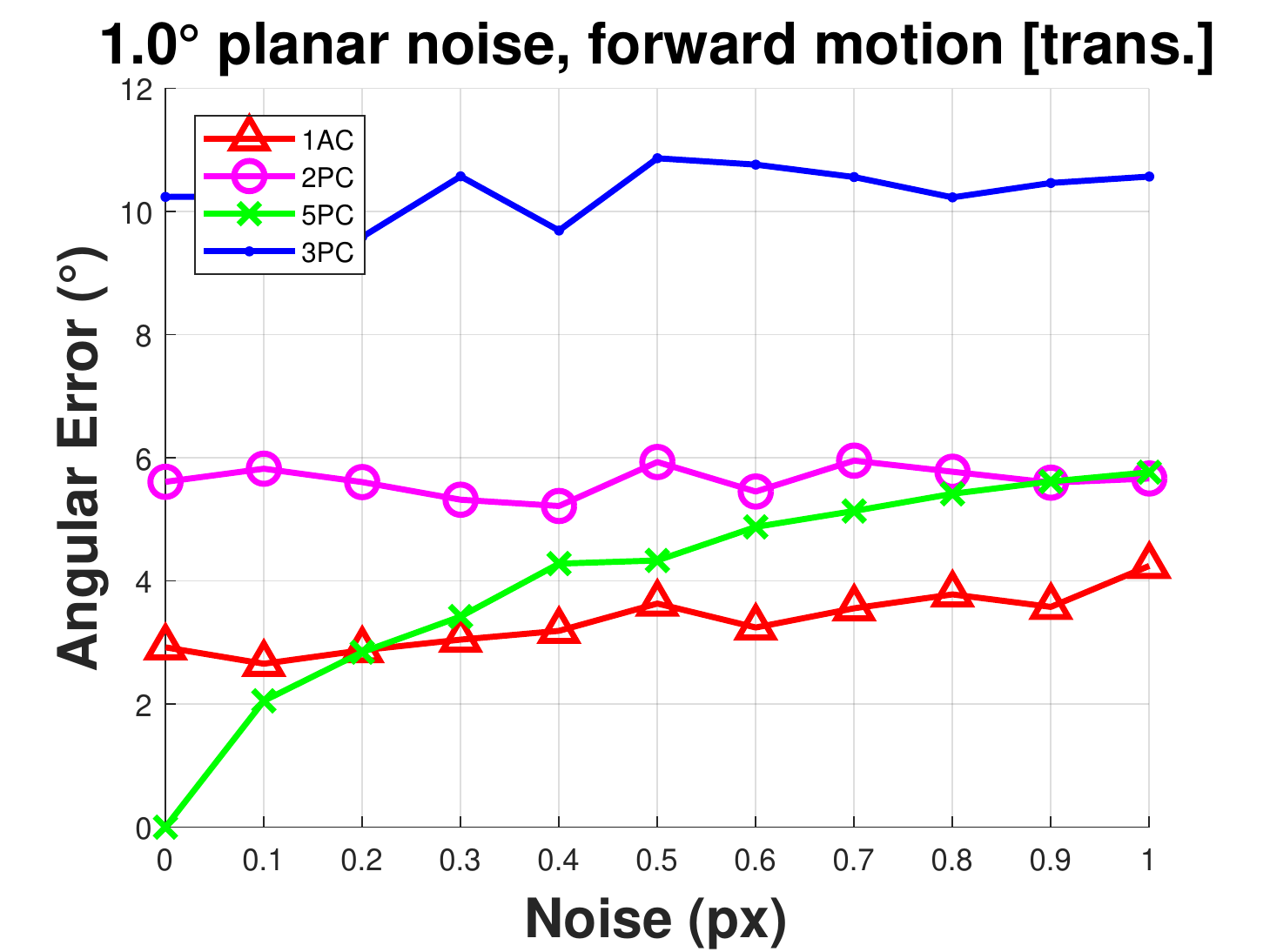}
            \includegraphics[width=0.245\columnwidth]{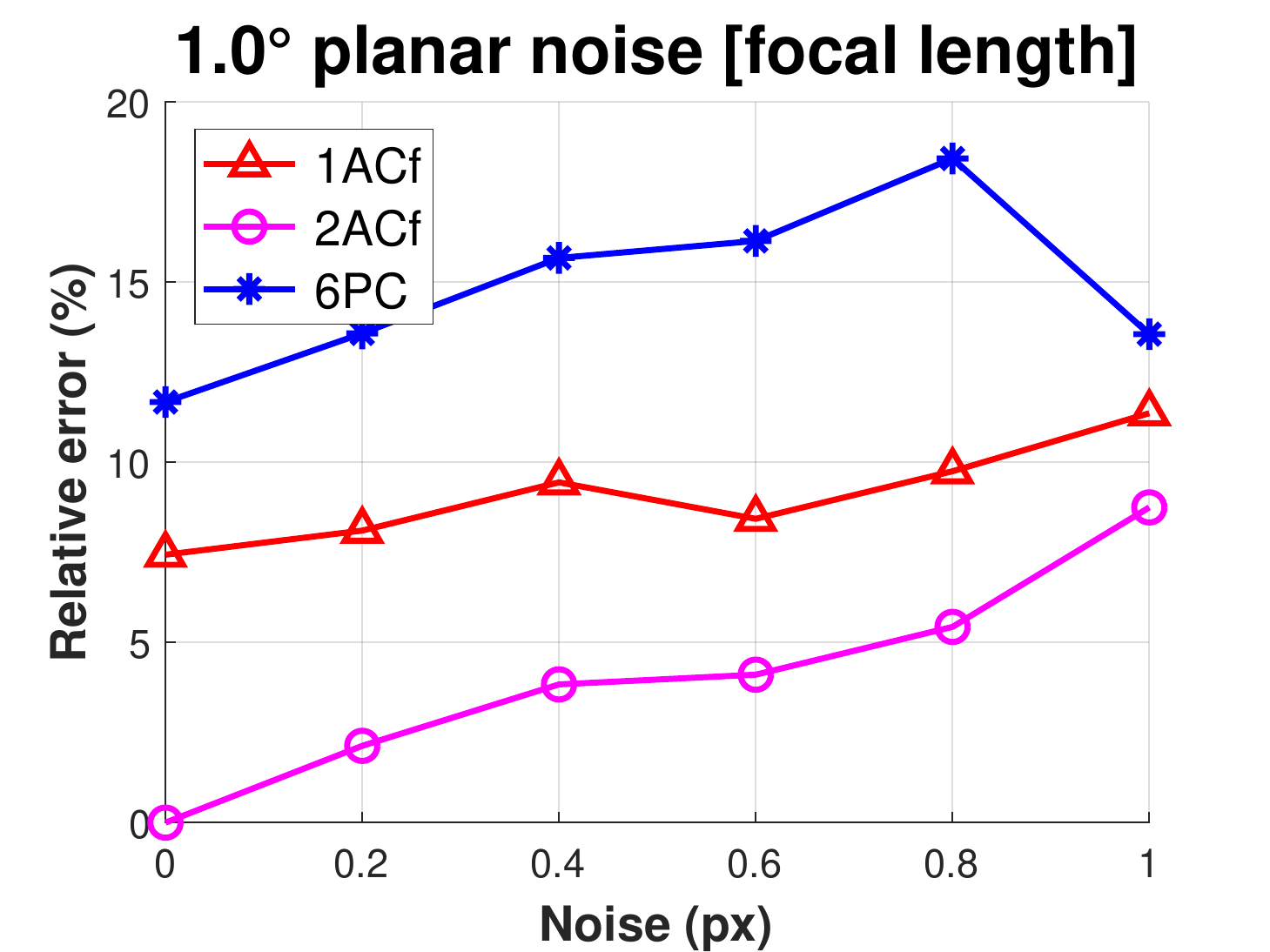}
    		\caption{General motion with corrupted planar constraint}
            \label{fig:synthesized_tests_bad_planar}
	\end{subfigure}
	
    \caption{(a) \textit{Calibrated case.} The mean angular error ($^\circ$) of the obtained rotations (left) and translations (right) is plotted as the function of the noise $\sigma$. The methods ran $1,000$ times for each $\sigma$. The compared methods are the proposed one (1AC), that of Choi et al.~\cite{choi2018} (2PC), David Nist\'er~\cite{nister2004efficient} (5PC) and the linear three-point algorithm (3PC). 
    For the first two plots, general planar motion was considered. For the last two ones, the cameras underwent a forward motion without rotation.  
    (b) \textit{Focal length estimation.}
    The compared methods are the proposed one (1ACf), the method of Barath et al.~\cite{barath2017minimal} (2ACf) and that of Hartley et al.~\cite{hartley2012efficient} (6PC).
    The frequencies ($1,000$ runs; vertical axis) of $\text{log}_{10}$ relative errors (horizontal) in the estimated focal lengths are shown in the 1st plot.  
    The 2nd one reports the relative focal length error (in $\%$) as the function of the noise $\sigma$. The mean angular errors ($^\circ$) of the obtained rotations (3rd plot) and translations (4th) are plotted as the function of the noise $\sigma$.
    (c) \textit{The planar constraint was corrupted} by rotating the vertical direction of the second camera by a random rotation with $\sigma = 1.0^\circ$. The first two plots show the errors in the rotations and translations in case of general planar motion. The 3rd plot reports the errors for forward motion. The last one shows the error of the estimated focal length. It can be seen that even though the corrupted planar constraint, the proposed 1AC and 1ACf methods work reasonably well. }
    \label{fig:synthesized_tests}
\end{figure*}
\vspace{-1.0ex}

The computational complexities of the competitor algorithms are compared in this section.
In Table~\ref{tab:computational_complexity}, each column shows the properties of a minimal solver. 
The first row consists of the major steps. 
For instance, $5 \times 9$ SVD + $10 \times 10$ Gauss-J + $10 \times 10$ EIG means that the steps are: the SVD decomposition of a $5 \times 9$ matrix, the Gauss-Jordan elimination of a $10 \times 10$ matrix and, finally, the eigendecomposition of a $10 \times 10$ matrix. 
In the second row, the implied computational complexities are summed. 
In the third one, the number of correspondences required for each solver is written. 
The fourth row lists example outlier ratios in the data. 
In the fifth one, the theoretical numbers of iterations of RANSAC~\cite{fischler1981random} are written for each outlier ratio with the required confidence set to $0.99$.
The last row shows the computational complexity of RANSAC combined with the minimal methods: it is the complexity of one iteration multiplied by the iteration number.
It can be seen that \textit{the proposed 1ACf and 1AC methods lead to the fewest iterations} and the smallest computational complexities. 
 
\subsection{Synthesized tests}

To test the proposed method and compare it to the state-of-the-art in a fully controlled environment, two perspective cameras were generated by their projection matrices $\Proj_1$ and $\Proj_2$. The cameras had common intrinsic parameters: $f_x = f_y = 600$ (focal length) and $[300 \quad 300]^\trans$ (principal point). 
In each test, a random plane was generated and 50 random points, from the plane, were projected into the cameras. Zero-mean Gaussian noise was added to the point coordinates with $\sigma$ standard deviation. To get the affine parameters for each point correspondence, the homography was estimated from the noisy correspondences. Then the noisy affine parameters for each correspondence were calculated from the corresponding homography~\cite{barath2016novel}. 

The error of an obtained rotation is the angle between vectors $\matr{R}_{\text{est}} \matr v$ and $\matr{R}_{\text{gt}} \matr v^\trans$ as follows: 
$
    \epsilon_\text{R} = \cos^{-1}((\matr{R}_{\text{est}} \matr v)^\trans (\matr{R}_{\text{gt}} \matr v)),
$
where $\matr v = [\frac{1}{\sqrt{3}}, \; \frac{1}{\sqrt{3}}, \; \frac{1}{\sqrt{3}}]^\trans$, $\matr{R}_{\text{est}}$ is the estimated and $\matr{R}_{\text{gt}}$ is the ground truth rotation matrix. The error of a translation is calculated in a fairly similar manner -- it is the angle between the ground truth and obtained translation vectors. The reported values are the averages over $1,000$ runs for every noise $\sigma$. 

\noindent \textit{Comparison of solvers for the calibrated case.}
For calibrated cameras, angles $\alpha$ and $\beta$ have to be estimated. The results are visualized in Fig.~\ref{fig:synthesized_tests_calibrated}. 
The competitor algorithms are the followings: (i) the proposed approach (1AC); (ii) the five-point method (5PC) proposed by David Nist\'er~\cite{Nister2003}; (iii) the two-point algorithm of \cite{choi2018} (2PC) and the linear three-point algorithm (3PC) described in Section~\ref{sec:cal_case}. For 5PC, the essential matrix is computed from five point correspondences assuming general 3D motion. 
3PC solves the linear system to which three point correspondences lead as it is described earlier.
%For solving the 2PC problem, we applied \cite{choi2018}. 
%
In Fig.~\ref{fig:synthesized_tests_calibrated}, the angular error (in degree) is plotted as the function of noise $\sigma$. 
General planar motion was considered for the first two plots showing the errors in the obtained rotations and translations. It can be seen that the 1AC is the second most accurate in both cases. However, the first one is different in these two cases. 
It was also investigated how the methods behave if the cameras undergo purely forward movement and the rotation around axis $\text{Y}$ is zero. See the last two plots of Fig.~\ref{fig:synthesized_tests_calibrated}.
In this case, 1AC obtains the most accurate rotations and the second most accurate translations. 

In the first three plots of Fig.~\ref{fig:synthesized_tests_bad_planar} we corrupted the planarity constraint by rotating the vertical direction of the second camera by a random angle with $\ang{1}$ standard deviation. 
Therefore, the problem has more DoF (i.e., three) than what planar methods consider. 
It can be seen that the proposed method is the most robust one to this kind of noise among the methods assuming planar movement, i.e.\ 2PC and 3PC. 
If the noise in the coordinates is $>0.2$ pixel, 1AC obtains the most accurate translations and the second most accurate rotations. If the noise is smaller than $0.2$ pixel, the general five point method~\cite{nister2004efficient} is the most accurate one. % as it can only cope with an estimation with $3$ DoF.

\noindent \textit{Comparison of solvers for the semi-calibrated case.}
In this test scenario, two motion angles and the common focal length of the two cameras are estimated. The results are shown in the second row of Fig.~\ref{fig:synthesized_tests_focal}. Three methods are compared: (i) the proposed approach (1ACf) explained in Section~\ref{sec:focal_solver}; (ii) the two-affine method~\cite{barath2017minimal} (2ACf) for general camera motion; (iii) and six-point technique of Hartley et al.~\cite{hartley2012efficient}. 
The left plot reports the numerical stability of the solvers: the frequencies (vertical axis; in $1,000$ runs) of the $\log_{10}$ relative errors (horizontal), i.e.\ $f_{\text{rel}} = (f_{\text{est}} - f_{\text{gt}}) / f_{\text{gt}}$, in the noise-free case. 
It can be seen that for methods 2ACf and 6PC, the frequency of errors around $10^{-2}$ -- $10^0$ is not zero. Therefore, there are cases when the desired focal length is not recovered by these methods even in the noise-free case. 
\textit{The proposed 1ACf is stable} with its peak over $10^{-9}$ relative error.
According to the other three plots of Fig.~\ref{fig:synthesized_tests_focal}, 1ACf obtains the second most accurate focal lengths slightly behind 2ACf; it leads to the most accurate rotation estimates and the recovered translations are the second best.  
The right plot of Fig.~\ref{fig:synthesized_tests_bad_planar}, i.e.\ when the planar constraint is corrupted, shows that 1ACf obtains the second best focal lengths even though the corrupted planarity. 

\subsection{Real-world experiments}

To test the proposed method on real-world data, we chose the Malaga\footnote{\url{https://www.mrpt.org/MalagaUrbanDataset}}~\cite{blanco2014malaga} dataset. This dataset was gathered entirely in urban scenarios with a car equipped with several sensors, including one high-resolution stereo camera and five laser scanners. We used the sequences of one camera and every $10$th image from each sequence. The methods were applied to every consecutive image pair of the $15$ sequences. Thus, they were tested on a total of $6,111$ image pairs. 
Finally, the stereo results are simply concatenated. 
To show solely the effect of the minimal solvers, each of them was applied to the consecutive frame pairs without optimizing the full path. 
%Thus the results were composed frame-by-frame without using a complex pipeline which provides an optimized path. 
The ground truth path was composed using the GPS coordinates provided in the dataset. 

\noindent \textit{Comparison of solvers for the calibrated case.}
We chose the same competitor methods as for the synthesized tests. Note that the one-point technique of Scaramuzza~\cite{Scaramuzza2011} could also be a competitor, however, it requires a special camera setting, i.e.\ the camera must be above the rear axle. This condition does not hold for this dataset. Affine correspondences are detected by Affine SIFT~\cite{yu2009fully} (ASIFT). The absolute length of the movement between each consecutive frame was computed from the given GPS coordinates. 
As a robust estimator, we chose histogram voting and Graph-Cut RANSAC~\cite{barath2017graph} since it can be considered as state-of-the-art and its implementation is publicly available\footnote{\url{https://github.com/danini/graph-cut-ransac}}. 
For GC-RANSAC, the required confidence was set to $0.99$ and the default setting, proposed by the authors, was used.

\begin{table}
\center
    \resizebox{0.99\columnwidth}{!}{\begin{tabular}{ l || c | c | c | c  }
    \hline
    	\cellcolor{black!20} & \cellcolor{black!20}\phantom{x}1AC + GC\phantom{x} & \cellcolor{black!20}\phantom{x}1AC + Hist\phantom{x} & \cellcolor{black!20}\phantom{x}3PC\phantom{x} & \cellcolor{black!20}\phantom{x}5PC\phantom{x} \\ 
    \hline
 	 	 Time (ms) & $749 \pm 672$ & $\textbf{16} \pm 10$ & \phantom{x}$598 \pm 530$\phantom{x} & \phantom{x}$588 \pm 527$\phantom{x} \\ 
 	 	 Sample \# & $\textbf{23} \pm 16$ & -- & $31 \pm 64$ & $38 \pm 32$ \\  
	\hline
    \end{tabular}}
	\caption{ Avg.\ and std.\ of the processing times (in milliseconds; in C++) and sample numbers for GC-RANSAC (2nd row) on the Malaga dataset. The average number of correspondences is $4,070$. The average processing time of the affine feature extraction by GPU-ASIFT~\cite{codreanu2013gpu} is $24$ ms on an NVIDIA GeForce GTX 980. }
\label{tab:processing_times}
\end{table}
\begin{figure}[h]
  	\centering
  	\includegraphics[width=0.97\columnwidth]{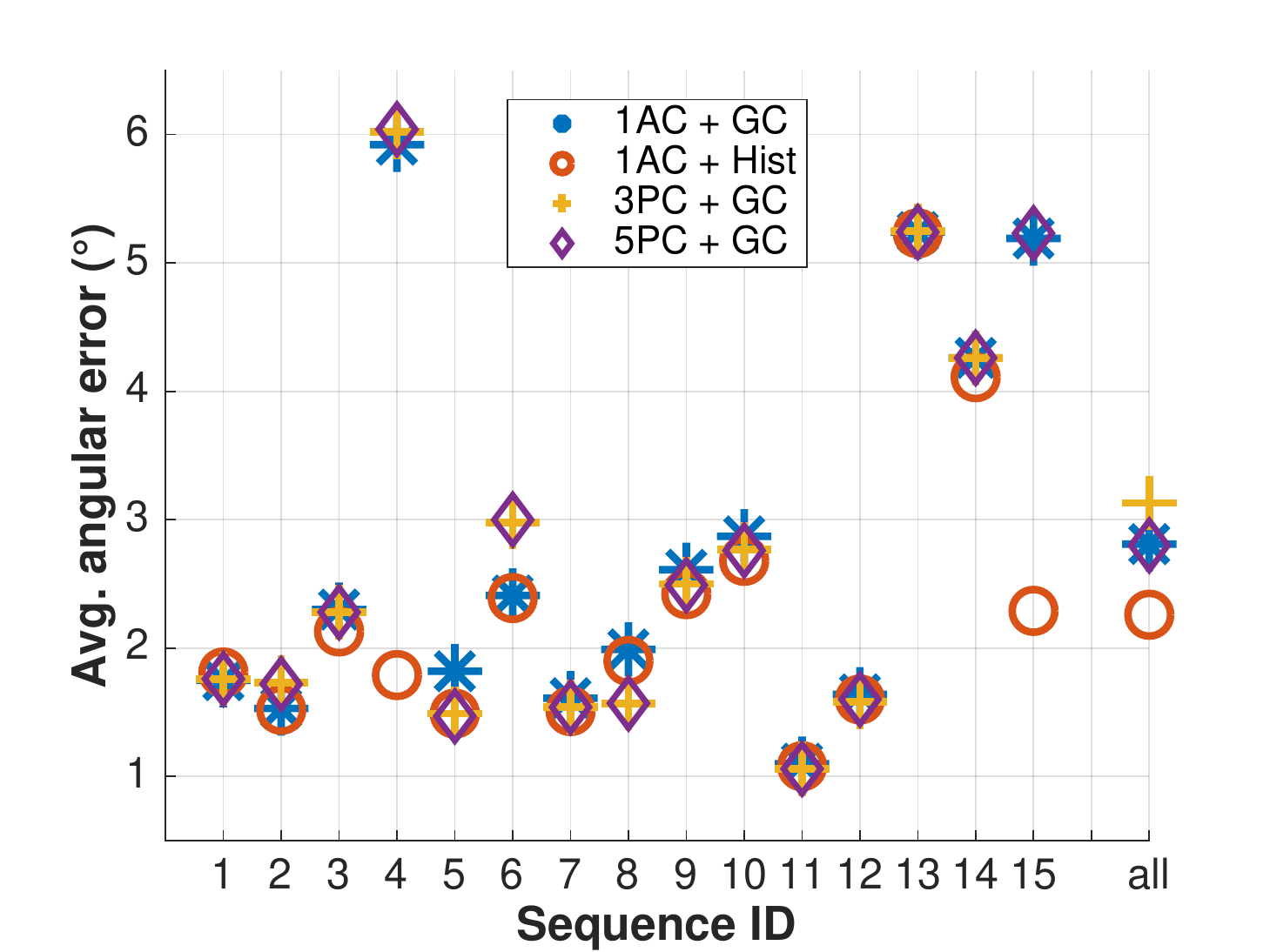}
  	\caption{ Avg.\ angular error (in degrees; vertical axis) on each sequence (horizontal) of the Malaga dataset. The compared methods are the proposed one (1AC), 3-point (3PC) and 5-point (5PC) algorithms. Suffix ``GC'' indicates that GC-RANSAC~\cite{barath2017graph} was applied as a robust estimator, while ``Hist'' stands for weighted histogram voting~\cite{bujnak2009robust}.
  	  The last column shows the average on all scenes ($6,111$ image pairs). }
	\label{fig:malaga_comparison}
\end{figure}

Figure~\ref{fig:malaga_comparison} reports the average angular error (in degrees; vertical axis) on each sequence (horizontal) of the Malaga dataset. In total, it was tested on $6,111$ image pairs. 
Example paths are in Fig.~\ref{fig:example_results}. 
It can be seen that, on average, the \textit{proposed method is superior} to the competitors algorithm in terms of geometric accuracy. 

Table~\ref{tab:processing_times} shows the processing time and the number of samples required by GC-RANSAC combined with each solver. 
GC-RANSAC with the proposed algorithm leads to the least number of samples, however, the processing time is slightly higher than that of the 3PC and 5PC algorithms. 
Using histogram voting as the robust estimator leads to an order of magnitude speedup. 
 
The feature extraction takes $24.07$ ms on an NVIDIA GeForce GTX 980 by our GPU-ASIFT~\cite{codreanu2013gpu} implementation for each image pair of size $1024 \times 768$.
The time of 1AC with histogram and that of the feature extraction takes $\approx40$ milliseconds which is \textit{far real time}.

\noindent \textit{Comparison of solvers for the semi-calibrated case.}
The accuracy of the estimated focal lengths was tested on the Malaga dataset. Due to having the intrinsic calibrations provided, the ground truth focal lengths are given. 
We applied the proposed 1ACf, the 2ACf~\cite{barath2017minimal} and 6PC~\cite{hartley2012efficient} methods to each consecutive image pair. 
Histogram voting was used for robust focal length estimation.
Figure~\ref{fig:real_focal_errors} reports the relative errors (vertical axis), for each sequence from the dataset (horizontal).
The relative error is $|f_{\text{est}} - f_{\text{gt}}| / f_{\text{gt}}$, where $f_{\text{est}}$ is the estimated and $f_{\text{gt}}$ is the ground truth focal length. 
In total, the algorithms were tested on $6,111$ image pairs. 
It can be seen, that the proposed 1ACf solver leads to the most accurate focal lengths on $11$ sequences out of the $15$. Also, the average error of 1ACf over all sequences is the lowest.

\section{Conclusion}

In this paper, the general relationship of affine correspondences and epipolar geometry was specialized considering that the stereo cameras move on a plane and the vertical camera directions are parallel.  
Exploiting the proposed relationship, two methods were proposed: one for estimating the relative camera motion from a single affine correspondence, another one to solve the semi-calibrated case, i.e.\ when the intrinsic calibration is known except the common focal length. 
As only one correspondence is minimally required, efficient robust estimators like histogram voting are applicable, leading to results superior to the state of the art both in terms of geometric accuracy and processing time. 

\newpage
\bibliographystyle{unsrt}
\bibliography{egbib}

\end{document}